\documentclass[10pt, a4paper]{article}
\pdfoutput=1
\usepackage{lrec}
\usepackage{multibib}
\newcites{languageresource}{Language Resources}
\usepackage{graphicx}
\usepackage{tabularx}
\usepackage{soul}

\usepackage{bbm}
\graphicspath{ {./images/} }
\usepackage{amsmath}
\usepackage{listings}
\usepackage{url}
\usepackage{xcolor}

\usepackage{titlesec}
\titleformat{\section}{\normalfont\large\bf\center}{\thesection.}{1em}{}
\titleformat{\subsection}{\normalfont\SmallTitleFont\bf\raggedright}{\thesubsection.}{1em}{}
\titleformat{\subsubsection}{\normalfont\normalsize\bf\raggedright}{\thesubsubsection.}{1em}{}
\renewcommand\thesection{\arabic{section}}
\renewcommand\thesubsection{\thesection.\arabic{subsection}}
\renewcommand\thesubsubsection{\thesubsection.\arabic{subsubsection}}

\usepackage{epstopdf}
\usepackage[utf8]{inputenc}

\usepackage{hyperref}
\DeclareUnicodeCharacter{F020}{~}
\usepackage{xstring}

\usepackage{color}

\title{TutorialVQA: Question Answering Dataset for Tutorial Videos}

\name{Anthony Colas$^*$, Seokhwan Kim$^\dagger$, Franck Dernoncourt$^\dagger$, \\Siddhesh Gupte$^*$, Daisy Zhe Wang$^*$, Doo Soon Kim$^\dagger$}

\address{ *University of Florida, $\dagger$ Adobe Research \\
         * 432 Newell Dr, Gainesville, FL 32611, $\dagger$ 345 Park Avenue
San Jose, CA 95110-2704 \\
         *\{acolas1, daisyw, siddhesh.gupte\}@ufl.edu\\
         $\dagger$\{seokim, dernonco,dkim\}@adobe.com\\}
\abstract{
Despite the number of currently available datasets on video-question answering, there still remains a need for a dataset involving multi-step and non-factoid answers. Moreover, relying on video transcripts remains an under-explored topic. To adequately address this, we propose a new question answering task on instructional videos, because of their verbose and narrative nature. While previous studies on video question answering have focused on generating a short text as an answer, given a question and video clip, our task aims to identify a span of a video segment as an answer which contains instructional details with various granularities. This work focuses on screencast tutorial videos pertaining to an image editing program. We introduce a dataset, TutorialVQA, consisting of about 6,000 manually collected triples of (video, question, answer span). We also provide experimental results with several baseline algorithms using the video transcripts. The results indicate that the task is challenging and call for the investigation of new algorithms. \\ \newline  }

\begin{document}

\maketitleabstract

\section{Introduction}

Video is the fastest growing medium to create and deliver information today.
Consequentially, videos have been increasingly used as main data sources in many question answering problems~\cite{yang2003videoqa,tapaswi2016movieqa,jang2017tgif,maharaj2017dataset,kim2017deepstory,jang2017tgif,zeng2017leveraging}.
These previous studies have mostly focused on factoid questions, each of which can be answered in a few words or phrases
generated by understanding multimodal contents in a short video clip.

However, this problem definition of video question answering causes some practical limitations for the following reasons.
First, factoid questions are just a small part of what people actually want to ask on video contents.
Especially if a short video is given to users, most fragmentary facts within the scope of previous tasks can be easily perceived by themselves even before asking questions.
Thus, video question answering is expected to provide answers to more complicated non-factoid questions beyond the simple facts. For example, users may have \textit{``how to"} type questions where the answer involves many fragmented steps to complete the task, as shown in Fig.~\ref{fig:ex}.

Accordingly, the answer format needs to also be improved towards more flexible ways besides multiple choice~\cite{tapaswi2016movieqa,jang2017tgif} or fill-in-the-blank questions~\cite{maharaj2017dataset,kim2017deepstory}.
Although open-ended video question answering~\cite{yang2003videoqa,jang2017tgif,zeng2017leveraging} has been explored, 
it still aims to generate just a short word or phrase-level answer, which is not enough to cover the various granularities of non-factoid question answering.

The other issue is that many videos with sufficient amounts of information, which a user is likely to pose questions on, have much longer lengths than the video clips in the existing datasets.
Therefore, in practice, the most relevant part of a whole video needs to be determined prior to each answer generation phase.
However, this localization task has been out of scope for previous studies.

\begin{figure*}[t]
\centering
\includegraphics[width=\textwidth]{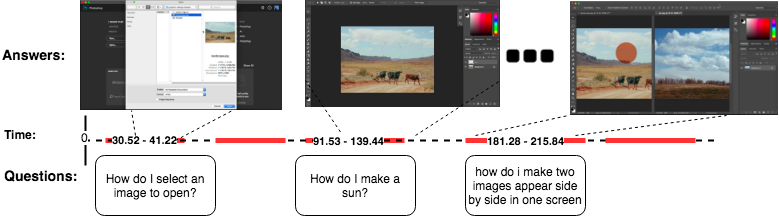}
\caption{An illustration of our task, where the red in the timeline indicates where answers can be found in a video.}
\label{fig:problem_definition}
\end{figure*}

In this work, we propose a new question answering problem for non-factoid questions on instructional videos. In accord with the nature of the media created for instructional purposes, we assume that many answers may already exist within the given video contents. Under this assumption, we formulate the problem as a localization task to specify the span of a video segment as the direct answer to a given video and question, as illustrated in Figure~\ref{fig:problem_definition}. 

The remainder of this paper is structured as follows: Section~\ref{sec:related} goes over some related work.
Section~\ref{sec:dataset} introduces TutorialVQA dataset as a case study of our proposed problem.
The dataset includes about 6,000 triples, comprised of videos, questions, and answer spans manually collected from screencast tutorial videos with spoken narratives for a photo-editing software. Section~\ref{sec:baseline} presents the baseline models and their experiment details on the sentence-level prediction and video segment retrieval tasks on our dataset. Then, we discuss the experimental results in Section~\ref{sec:discussion} and conclude the paper in Section~\ref{sec:conclusion}.

\section{Related Work}
\label{sec:related}
Most relevant to our proposed work is the reading comprehension task, which is a question answering task involving a piece of text such as a paragraph or article. Such datasets for the reading comprehension task, such as SQuAD \cite{rajpurkar2016squad} based on Wikipedia, TriviaQA \cite{joshi2017triviaqa} constructed from trivia questions with answer evidence from Wikipedia, or those from Hermann et al. based on CNN and Daily Mail articles \cite{hermann2015teaching} are factoid-based, meaning the answers typically involve a single entity. Differing from video transcripts, the structures of these data sources, namely paragraphs from Wikipedia and news sources, are typically straightforward since they are meant to be read. In contrast, video transcripts originate from spoken dialogue, which can be verbose, unstructured, and disconnected. Furthermore, the answers in instructional video transcripts can be longer, spanning multiple sentences if the process is multi-step or even fragmented into multiple segments throughout the video.

Visual corpora in particular have proven extremely valuable to visual questions-answering tasks \cite{antol2015vqa}, the most similar being MovieQA \cite{tapaswi2016movieqa} and VideoQA \cite{yang2003videoqa}. Similar to how our data is generated from video tutorials, the MovieQA and VideoQA corpus is generated from movie scripts and news transcipts, respectively. MovieQA's answers have a shorter span than the answers collected in our corpus, because questions and answer pairs were generated after each paragraph in a movie's plot synopsis \cite{tapaswi2016movieqa}. The MovieQA dataset also contains specific annotated answers with incorrect examples for each question. In the VideoQA dataset, questions focus on a single entity, contrary to our instructional video dataset. Although not necessarily a visual question-answering task, the work proposed by Gupta et al. involved answering questions over transcript data \cite{Gupta2018NeuralAR}. Contrary to our work, Gupta et al.'s dataset is not publically available and their examples only showcase factoid-style questions involving single entity answers.

Malmaud et al. focus on aligning a set of instructions to a video of someone carrying out those instructions \cite{malmaud2015s}. In their task, they use the video transcript to represent the video, which they later augment with a visual cue detector on food entities. Their task focuses on procedure-based cooking videos, and contrary to our task is primarily a text alignment task. In our task we aim to answer questions\textemdash using the transcripts\textemdash on instructional-style videos, in which the answer can involve steps not mentioned in the question.

\section{TutorialVQA Dataset}
\label{sec:dataset}
In this section, we introduce the TutorialVQA dataset and describe the data collection process.\footnote{https://github.com/acolas1/TutorialVQAData} 

\subsection{Overview}
Our dataset consists of 76 tutorial videos pertaining to an image editing software. 
All of the videos include spoken instructions which are transcribed and manually segmented into multiple segments. Specifically, we asked the annotators to manually divide each video into multiple segments such that each of the segments can serve as an answer to any question. For example, Fig.~\ref{fig:problem_definition} shows example segments marked in red (each which are a complete unit as an answer span). Each sentence is associated with the starting and ending time-stamps, which can be used to access the relevant visual information. 

\begin{figure}[t]
\centering
\includegraphics[width=8cm]{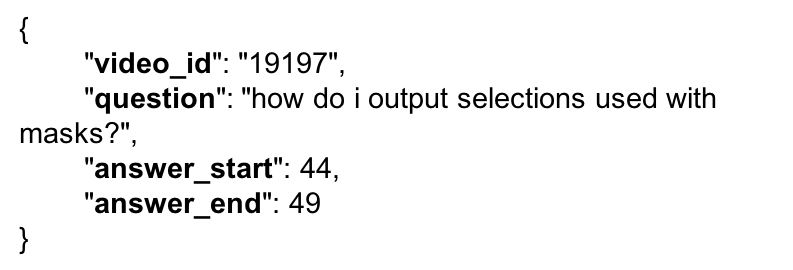}
\caption{An example of a QA annotation.}
\label{fig:ex}
\end{figure}

The dataset contains 6,195 non-factoid QA pairs, where the answers are the segments that were manually annotated. Fig.~\ref{fig:ex} shows an example of the annotations. \texttt{video\_id} can be used to retrieve the video information such as meta information and the transcripts. \texttt{answer\_start} and \texttt{answer\_end} denote the starting and ending sentence indexes of the answer span. Table~\ref{tab:data} shows the statistics of our dataset, with each answer segment having on average about 6 sentences, showing that our answers are more verbose than those in previous factoid QA tasks.
\begin{table}[t]
\begin{tabular}{l c r}
number of videos & & 76           \\ number of segments & & 408           \\
number of QA pairs & & 6,195        \\ 
avg. length of answer (sec) & & 31.39\\ 
avg. length of transcript (sentences) & & 48 \\ 
avg. length of question (words)  & & 9 \\ 
avg. length of answer (sentences) & & 6 \\ 
\end{tabular}
\caption{Statistics of TutorialVQA dataset.}
\label{tab:data}
\end{table}

\subsection{Basis}
We chose videos pertaining to an image editing software because of the complexity and variety of tasks involved. In these videos, a narrator is communicating an overall goal by utilizing an example. For example, in ~\ref{fig:problem_definition} the video pertains to combining multiple layers into one image. However, throughout the videos multiple subtasks are achieved, such as the opening of multiple images, the masking of images, and the placement of two images side-by-side. These subtasks involve multiple steps and are of interest to us in segmenting the videos. Each segment can be seen as a subtask within a larger video dictating an example. We thus chose these videos because of the amount of procedural information stored in each video for which the user may ask. Though there is only one domain, each video corresponds to a different overall goal.

\subsection{Data Collection}
We downloaded 76 videos from a tutorial website about an image editing program.\footnote{https://helpx.adobe.com/photoshop/tutorials.html} Each video is pre-processed to provide the transcripts and the time-stamp information for each sentence in the transcript. We then used Amazon Mechanical Turk\footnote{https://www.mturk.com/} to collect the question-answer pairs.\footnote{We recruited workers who have completed at least 100 AMT tasks and have at least a 95\% approval rating.} One naive way of collecting the data is to prepare a question list and then, for each question, ask the workers to find the relevant parts in the video. However, this approach is not feasible and error-prone because the videos are typically long and finding a relevant part from a long video is difficult. Doing so might also cause us to miss questions which were relevant to the video segment. Instead, we took a reversed approach. First, for each video, we manually identified the sentence spans that can serve as answers. These candidates are of various granularity and may overlap. The segments are also complete in that they encompass the beginning and end of a task. In total, we identified 408 segments from the 76 videos. Second we asked AMT workers to provide question annotations for the videos.

Our AMT experiment consisted of two parts. In the first part, we presented the workers with the video content of a segment. For each segment, we asked workers to generate questions that can be answered by the presented segment. We did not limit the number of questions a worker can input to a corresponding segment and encouraged them to input a diverse set of questions which the span can answer. Along with the questions, the workers were also required to provide a justification as to why they made their questions. We manually checked this justification to filter out the questions with poor quality by removing those questions which were unrelated to the video. One initial challenge worth mentioning is that at first some workers input questions they had about the video and not questions which the video could answer. This was solved by providing them with an unrelated example. The second part of the question collection framework consisted of a paraphrasing task. In this task we presented workers with the questions generated by the first task and asked them to write the questions differently while keeping the semantics the same. In this way, we expanded our question dataset. After filtering out the questions with low quality, we collected a total of 6,195 questions. 

It is important to note the differences between our data collection process and the the query generation process employed in the Search and Hyperlinking Task at MediaEval \cite{eskevich2014search}. In the Search and Hyperlinking Task, 30 users were tasked to first browse the collection of videos, select interesting segments with start and end times, and then asked to conjecture questions that they would use on a search query to find the interesting video segments. This was done in order to emulate their thought process mechanism. While the nature of their task involves queries relating to the overall videos themselves, hence coming from a video's interestingness, our task involves users already being given a video and formulating questions where the answers themselves come from within a video. By presenting the same video segment to many users, we maintain a consistent set of video segments and extend the possibility to generate a diverse set of question for the same segment. 
\subsection{Dataset Details}
Table ~\ref{tab:questions} presents some extracted sample questions from our dataset. The first column corresponds to an AMT generated question, while the second column corresponds to the video ID where the segment can be found. As can be seen in the first two rows, multiple types of questions can be answered within the same video (but different segments). The last two rows display questions which belong to the same segment but correspond to different properties of the same entity, `crop tool'. Here we observe different types of questions, such as ``why", ``how", ``what", and ``where", and can see why the answers may involve multiple steps. Some questions that the worked paraphrased were in the ``yes/no" style, however our answer segments then provide an explanation to these questions. 

Each answer segment was extracted from an image editing tutorial video that involved multiple steps and procedures to produce a final image, which can partially be seen in ~\ref{fig:problem_definition}. The average number of sentences per video was approximately 52, with the maximum number of sentences contained in a video being 187. The sub-tasks in the tutorial include segments (and thus answers) on editing parts of images, instructions on using certain tools, possible actions that can be performed on an image, and identifying the locations of tools and features, with the shortest and longest segment having a span of 1 and 37 sentences respectively, demonstrating the heterogeneity of the answer spans.

\begin{table}[]
\centering
\begin{tabular}{|l|l|}
\hline
\textbf{Example}                                                                                  & \textbf{Video ID} \\ \hline
\begin{tabular}[c]{@{}l@{}}why might some alignment options\\ appear grey?\end{tabular}           & 4051              \\ \hline
\begin{tabular}[c]{@{}l@{}}how would i go about selecting \\ all the layers at once?\end{tabular} & 4051              \\ \hline
what does the crop tool do?                                                                       & 4177              \\ \hline
where is the crop tool located?                                                                   & 4177              \\ \hline
\end{tabular}
\caption{Examples of question variations}
\label{tab:questions}
\end{table}

\section{Baselines}
\label{sec:baseline}
Our video question answering task is novel and to our knowledge, no model has been designed specifically for this task. As a first step towards solving this problem, we evaluated the performance of state-of-the-art models developed for other QA tasks, including a sentence-level prediction task and two segment retrieval tasks. In this section, we report their results on the TutorialVQA dataset.\footnote{For the baselines, we considered only the transcript information, since it was non-trivial to include the other modalities such as video. In the future we plan to develop a multi-modal approach.}

\subsection{First Baseline: Sentence-level prediction}
\begin{figure*}[t]
\centering
\includegraphics[width=12cm]{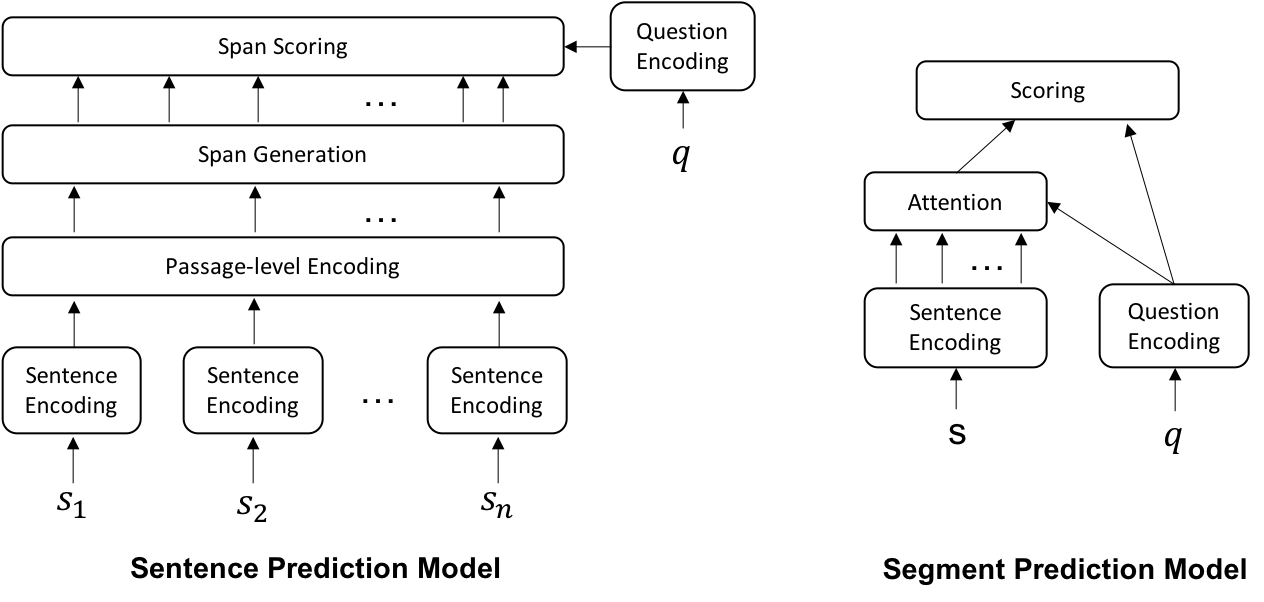}
\caption{Baseline models for sentence-level prediction and video segment retrieval tasks.}
\label{fig:models}
\end{figure*}

Given a transcript (a sequence of sentences) and a question, the first baseline predicts \textit{(starting sentence index, ending sentence index)}. The model is based on RaSor \cite{lee2016learning}, which has been developed for the SQuAD QA task \cite{rajpurkar2016squad}. RaSor concatenates the embedding vectors of the starting and the ending words to represent a span. Following this idea, the first baseline represents a span of sentences by concatenating the vectors of the starting and ending sentences. The left diagram in Fig.~\ref{fig:models} illustrates the first baseline's model. 

\textbf{Model.} The model takes two inputs, a transcript, $\{s_1, s_2, ... s_n\}$ where $s_i$ are individual sentences and a question, $q$. The output is the span scores, $y$, the scores over all possible spans. GLoVe \cite{pennington2014glove} is used for the word representations in the transcript and the questions. We use two bi-LSTMs~\cite{bilstm} to encode the transcript. 
\begin{align*}
h_i &= biLSTM_{last}(s_i) \mbox{ for i = 1..n} & \mbox{(Sent Encoding)}\\
p_i &= biLSTM_{all}(\{h_1, .., h_n\}) & \mbox{(Psg Encoding)}
\end{align*}

where n is the number of sentences.\footnote{$biLSTM_{last}$ produces only the last hidden vector while $biLSTM_{all}$ produces all hidden vectors along the sequence.} The output of Passage-level Encoding, $p$, is a sequence of vector, $p_i$, which represents the latent meaning of each sentence. Then, the model combines each pair of sentence embeddings ($p_i$, $p_j$) to generate a span embedding.
\begin{align*}
r_{ij} &= [p_i, p_j] \mbox{ for i, j = 1..n} & \mbox{ (Span Generation)} 
\end{align*}
where [$\cdot$,$\cdot$] indicates the concatenation. Finally, we use a one-layer feed forward network to compute a score between each span and a question. 
\begin{align*}
h^q &= biLSTM_{last}(q) & \mbox{(Span Scoring)}\\
y_{ij} &= \mbox{Softmax(FFN}([r_{ij}, h^q])) & 
\end{align*}
  
In training, we use cross-entropy as an objective function. In testing, the span with the highest score is picked as an answer. 

\textbf{Metrics.} We use tolerance accuracy ~\cite{tsunoo2017hierarchical}, which measures how far away the predicted span is from the gold standard span, as a metric. The rationale behind the metric is that, in practice, it suffices to recommend a rough span which contains the answer -- a difference of a few seconds would not matter much to the user. 

Specifically, the predicted span is counted as correct if $|pred_{start} - gt_{start}| + |pred_{end} - gt_{end}| <=$ $k$, where $pred_{start/end}$ and $gt_{start/end}$ indicate the indices of the predicted and ground-truth starting and ending sentences, respectively. We then measure the percentage of correctly predicted questions among the entire test questions. 

\subsection{Second baseline: Segment retrieval}
\label{sec:baseline2}
We also considered a simpler task by casting our problem as a retrieval task. Specifically, in addition to a plain transcript, we also provided the model with the segmentation information which was created during the data collection phrase (See Section.~\ref{sec:dataset}). Note that each segments corresponds to a candidate answer. Then, the task is to pick the best segment for given a query. This task is easier than the first baseline's task in that the segmentation information is provided to the model. Unlike the first baseline, however, it is unable to return an answer span at various granularities. The second baseline is based on the attentive LSTM~\cite{tan2016improved}, which has been developed for the InsuranceQA task. The right diagram in Fig.~\ref{fig:models} illustrates the second baseline's model. 

\textbf{Model.} The two inputs, $s$ and $q$ represent the segment text and a question. The model first encodes the two inputs. 
\begin{align*}
h^s &= biLSTM_{all}(s) & \mbox{(Sentence Encoding)}\\
h^q &= biLSTM_{last}(q) & \mbox{(Question Encoding)}
\end{align*}

$h^s$ is then re-weighted using attention weights. 
\begin{align*}
a &= FFN([h^s, h^q]) & \mbox{(Attention)} \\
h'^s & =a \odot h^s
\end{align*}

where $\odot$ denotes the element-wise multiplication operation. The final score is computed using a one-layer feed-forward network. 
\begin{align*}
y &= Softmax(FFN([h_s, h_q])) & \mbox{(Scoring)}
\end{align*}

During training, the model requires negative samples. For each positive example, (question, ground-truth segment), all the other segments in the same transcript are used as negative samples. Cross entropy is used as an objective function.

\textbf{Metrics.} We used accuracy and MRR (Mean Reciprocal Ranking) as metrics. The accuracy is 
\begin{align*}
\frac{\mbox{\# of questions where the top answer is correct}}{\mbox{the total \# of questions}}
\end{align*}

\medskip

We split the ground-truth dataset to train/dev/test into the ratio of 6/2/2. The resulting size is 3,718 (train), 1,238 (dev) and 1,239 QA pairs (test).

\subsection{Third baseline: Pipeline Segment retrieval}
We construct a pipelined approach through another segment retrieval task, calculating the cosine similarities between the segment and question embeddings. In this task however, we want to test the accuracy of retrieving the segments given that we first retrieve the correct video from our 76 videos. First, we generate the TF-IDF embeddings for the whole video transcripts and questions. The next step involves retrieving the videos which have the lowest cosine distance between the video transcripts and question. We then filter and store the top ten videos, reducing the number of computations required in the next step. Finally, we calculate the cosine distances between the question and the segments which belong to the filtered top 10 videos, marking it as correct if found in these videos. While the task is less computationally expensive than the previous baseline, we do not learn the segment representations, as this task is a simple retrieval task based on TF-IDF embeddings.\\
\textbf{Model.} The first two inputs are are the question, \textit{q}, and video transcript, \textit{v}, encoded by their TF-IDF vectors: \cite{tata2007estimating}:
\begin{align*}
v_{TF-IDF} &= TF-IDF(v) & \mbox{(Video Encoding)}\\
q_{TF-IDF} &= TF-IDF(q) & \mbox{(Question Encoding)} 
\end{align*}
We then filter the top 10 video transcripts(out of 76) with the minimum cosine distance, and further compute the TF-IDF vectors for their segments, S\textsubscript{top10}\textsuperscript{n}, where \textit{n} = 10. We repeat the process for the corresponding segments:
\begin{align*}
s_{TF-IDF} &= TF-IDF(s) & \mbox{(Segment Encoding)}\\
\end{align*}
selecting the segment with the minimal cosine distance distance to the query. \\
\textbf{Metrics.} To evaluate our pipeline approach we use overall accuracy after filtering and accuracy given that the segment is in the top 10 videos. While the first metric is similar to ~\ref{sec:baseline2}, the second can indicate if initially searching on the video space can be used to improve our selection:
\begin{align*}
\frac{\mbox{\# of questions where the top answer is correct}}{\mbox{the total \# of questions answerable by top 10 videos}}
\end{align*}

\subsection{Results}

Tables~\ref{tab:eval1}, ~\ref{tab:eval2}, ~\ref{tab:eval3} show the results. First, the tables show that the two first baselines under-perform for our task. Even with a tolerance window of 6, the first baseline merely achieves an accuracy of .14. The second baseline, despite being a simpler task, has only an accuracy of .23. Second, while we originally hypothesized that the segment selection task should be easier than the sentence prediction task, Table~\ref{tab:eval2} shows that the task is also challenging. One possible reason is that the segments contained within the same transcript have similar contents, due to the composition of the overall task in each video, and differentiating among them may require a more sophisticated model than just using a sequence model for segment representation. Table ~\ref{tab:eval3} shows the accuracy of retrieving the correct segment, for baseline both overall and given that the video selected is within the top 10 videos. While the overall accuracy is only .16, by reducing the search space to 10 relevant videos our accuracy increases to .6385. In future iterations, it may then be useful to find better approaches in filtering large paragraphs of text before predicting the correct segment.

\begin{table}[t]
\centering
\begin{tabular}{c c c c c}
$k$ & & Train & Dev & Test\\ [0.5ex]
\hline
0 & & .0506& .0541 & .0523\\
2 & & .0825& .0645 & .0667\\
4 & & .1143& .1129 & .1133\\
6 & & .1412 & .1348 & .1443\\
\end{tabular}
\caption{Sentence-level prediction results for the first baseline with different tolerance window sizes $k$.}
\label{tab:eval1}
\end{table}

\begin{table}[t]
\centering
\begin{tabular}{l c c }
Model & MRR& ACC\\ [0.5ex]
\hline
Attentive LSTM & .4689 & .2341\\
\end{tabular}
\caption{Video segment retrieval results for the second baseline}
\label{tab:eval2}
\end{table}

\begin{table}[t]
\centering
\begin{tabular}{c c}
Overall & Given the video is in top 10 \\ [0.5ex] 
\hline
.1579 & .6385                             
\end{tabular}
\caption{Segment level prediction for the third baseline, both overall and given that the video is in the top 10.}
\label{tab:eval3}
\end{table}

\section{Discussion and Future Work}
\label{sec:discussion}
We performed an error analysis on the first baseline's results. We first observe that, in 92\% of the errors, the predicted span and the ground-truth overlap. Furthermore, in 56\% of the errors, the predicted spans are a subset or superset of the ground-truth spans. This indicates that the model finds the rough answer regions but fails to locate the precise boundaries. To address this issue, we plan on exploring the Pointer-network ~\cite{vinyals2015pointer}, which finds an answer span by selecting the boundary sentences. Unlike the first baseline which avoids an explicit segmentation step, the Pointer-network can explicitly model which sentences are likely to be a boundary sentence. Moreover, the search space of the spans in the Pointer-network is $2n$ where $n$ is the number of sentences, because it selects only two boundary sentences. Note that the search space of the first baseline is $n^2$. A much smaller search space might improve the accuracy by making the model consider fewer candidates.

In future work, we also plan to use multi-modal information. While our baselines only used the transcript, complementing the narratives with the visual information may improve the performance, similarly to the text alignment task in \cite{malmaud2015s}. 

\section{Conclusion}
\label{sec:conclusion}

We have described the collection, analysis,
and baseline results of TutorialVQA, a new type
of dataset used to find answer spans in tutorial
videos. Our data collection method for question-answer pairs on instructional video can be further adopted to other domains where the answers involve multiple steps and are part of an overall goal, such as cooking or educational videos. We have shown that current baseline models
for finding the answer spans are not sufficient
for achieving high accuracy and hope that by releasing
this new dataset and task, more appropriate
question answering models can be developed
for question answering on instructional videos.

\section{Bibliographical References}
\label{main:ref}

\bibliographystyle{lrec}
\bibliography{lrec2020W-xample-kc}


\end{document}